\documentclass[runningheads,a4paper]{llncs}

\usepackage{amssymb}
\usepackage{graphicx}

\usepackage{color}
\usepackage{times,amsmath} % Add all your packages here
\usepackage{psfrag}
\usepackage{stfloats}
\usepackage{multirow}
\usepackage{rotating}
\usepackage{amssymb}
\usepackage{paralist}
\usepackage{caption}
\usepackage{floatflt}
\usepackage{wrapfig}

\newcommand{\bmx}[0]{\begin{bmatrix}}
\newcommand{\emx}[0]{\end{bmatrix}}

\newcommand{\qlay}[1]{\left[#1\right]}

\newcommand{\matr}[1]{\mathbf{#1}}

\newcommand{\mW}[0]{\matr{W}}

\newcommand{\E}[0]{\mathbb{E}}

\usepackage{url}
\urldef{\mails}\path|{firstname.lastname@aalto.fi}|
\newcommand{\keywords}[1]{\par\addvspace\baselineskip
\noindent\keywordname\enspace\ignorespaces#1}

\begin{document}

\mainmatter  % start of an individual contribution

\title{Understanding Dropout: \\Training Multi-Layer Perceptrons \\ with Auxiliary
Independent Stochastic Neurons}
\titlerunning{Understanding Dropout with Auxiliary
Independent Stochastic Neurons}

%\titlerunning{Modified Gaussian-Bernoulli Restricted
%Boltzmann Machines}

%\author{\textit{arbitrary order}, KyungHyun Cho and Xi Chen}
\author{KyungHyun Cho}%

\institute{Department of Information and Computer Science\\
Aalto University School of Science, Finland\\
\mails}


%\toctitle{Lecture Notes in Computer Science}
%\tocauthor{Authors' Instructions}
\maketitle

\begin{abstract}
    In this paper, a simple, general method of adding
    auxiliary stochastic neurons to a multi-layer perceptron
    is proposed. It is shown that the proposed method is a
    generalization of recently successful methods of dropout
    \cite{Hinton2012}, explicit noise injection
    \cite{Vincent2010,Bishop1995} and semantic hashing
    \cite{Salakhutdinov2009s}. Under the proposed framework,
    an extension of dropout which allows using separate
    dropping probabilities for different hidden neurons, or
    layers, is found to be available. The use of different
    dropping probabilities for hidden layers separately is
    empirically investigated.

\end{abstract}

\section{Introduction}

In this paper, we describe a simple extension to a
multi-layer perceptron (MLP) that unifies some of the
recently proposed training tricks for training an MLP. For
example, the proposed extension is a generalization of using
dropout for training an MLP \cite{Hinton2012}.

The proposed method extends a conventional, deterministic
MLP by augmenting each hidden neuron with an auxiliary
stochastic neuron of which activation needs to be sampled.
The activation of the added stochastic neurons is
independent of all other variables in the MLP, and the
weight of the edge connecting from the auxiliary neuron to
the existing hidden neuron is fixed and not learned.
Consequently, learning the parameters of the extended MLP
does not require any special learning algorithm but can use
a standard backpropagation \cite{Rumelhart1986}.

This paper starts by briefly describing the proposed method
of adding auxiliary stochastic neurons to an MLP. Then, it
is described how dropout \cite{Hinton2012} and explicit
noise injection \cite{Vincent2010,Bishop1995} as well as
semantic hashing \cite{Salakhutdinov2009s} are all special
cases of the proposed framework. Understanding the method of
dropout under the proposed framework reveals that it is
possible to use separate dropping probabilities for hidden
neurons in a single MLP, and empirical investigation is
provided on using different dropping probabilities for
separate hidden layers.

\section{Perceptron with Auxiliary Stochastic Neuron}

For each hidden neuron $h_j^{\qlay{l}}$ in the $l$-th hidden
layer, we introduce an independent stochastic neuron
$r_j^{\qlay{l}}$ connected to $h_j^{\qlay{l}}$ with the edge
weight $u_j^{\qlay{l}}$. The edge weight $u_j^{\qlay{l}}$ is
\textit{not} learned but fixed to a certain constant either
indefinitely or for each forward computation\footnote{
The author acknowledges that a similar method of a hidden
neuron having an
independent noise source, called a \textit{semi-hard
stochastic neuron}, has
been recently proposed in \cite{Bengio2013sto} independently
of this work.
}.

%\begin{figure}[h]
\begin{wrapfigure}{I}{0.4\textwidth}
    \psfrag{h1}[Bl][Bl][1.0][0]{$h_{i-1}^{\qlay{l-1}}$}
    \psfrag{h2}[Bc][Br][1.0][0]{$h_{i}^{\qlay{l-1}}$}
    \psfrag{h3}[Bl][Br][1.0][0]{$h_{i+1}^{\qlay{l-1}}$}
    \psfrag{h}[Bl][Bl][1.0][0]{$h_{j}^{\qlay{l}}$}
    \psfrag{r}[Bl][Tl][1.0][0]{$r_{j}^{\qlay{l}}$}
    \psfrag{v}[Br][Br][1.0][0]{$\mW^{\qlay{l}}$}
    \psfrag{w}[Br][Br][1.0][0]{$\mW^{\qlay{l-1}}$}
    \psfrag{u}[Bl][Bl][1.0][0]{$u_j^{\qlay{l}}$}
    \centering
    \includegraphics[width=0.3\textwidth]{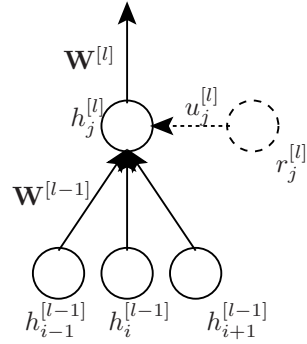}
    \caption{Illustration of adding an auxiliary stochastic
    hidden neuron (marked by a dashed circle).}
    \label{fig:aux_general}
    \vspace{-3mm}
\end{wrapfigure}
%\end{figure}

The auxiliary stochastic neuron $r_j^{\qlay{l}}$ follows a
predefined probability distribution, and its value is sampled
at each evaluation of $h_j^{\qlay{l}}$. Since there is no
incoming edge to the auxiliary neuron, the neuron is
independent of any other variable in the MLP. In this case,
the activation of the $j$-th hidden neuron in the $l$-th
layer is
\begin{align*}
    h_j^{\qlay{l}} = \phi\left( \sum_{i} h_i^{\qlay{l-1}}
    w_{ij}^{\qlay{l-1}} + r_j^{\qlay{l}} u_j^{\qlay{l}} \right),
\end{align*}
where $\phi$ is a nonlinear function. $h_i^{\qlay{l-1}}$ and
$w_{ij}^{\qlay{l-1}}$ are the $i$-th hidden neuron in the
$(l-1)$-th hidden layer and the edge weight between
$h_i^{\qlay{l-1}}$ and $h_j^{\qlay{l}}$, respectively. A hyperbolic tangent
function $\tanh(\alpha)$ or a rectified linear function
$\max(0, \alpha)$ is a common choice. See
Fig.~\ref{fig:aux_general} for the illustration.

\subsection{Learning and Prediction}

It is straightforward to learn the parameters of this MLP.
Since we do not attempt to learn $u_j^{\qlay{l}}$, a usual
backpropagation \cite{Rumelhart1986} can be used. Only
difference from the ordinary MLP which does not have
auxiliary stochastic neurons is that the activations of the
auxiliary neurons need to be sampled during the forward
computation. 

However, with a fixed set of parameters, either learned or
predetermined by a user, it is not trivial to make a
prediction given a new sample. Due to the stochastic
activation of the auxiliary neurons, each forward
computation of the output neurons will differ. A most
obvious approach is to compute the output activation several
times, and take the average or pick the most frequent one.
This is however not preferred due to the increased
computational cost as well as potentially high variance. 

Another, more preferred way is to compute the expected
activation of the output neurons over the distribution
defined by the auxiliary stochastic neurons. This is often
difficult as well due to the use of nonlinear activation
functions. However, it is possible to linearize the
computational path by approximating each nonlinear function
linearly and push down the expectation operator to each
auxiliary neuron.  One can, then, compute and use the
approximate expected activation of the output neurons as the
final prediction.

\section{Understanding Dropout with Auxiliary Stochastic
Neurons}
\label{sec:dropout}

A dropout is a regularization technique which forces the
activations of a randomly selected half of hidden neurons in
each layer to zero when training an MLP. Just like the
proposed method of adding auxiliary stochastic neurons
training an MLP with dropout changes the forward computation
only and leaves the error backpropagation as it is.

\subsection{Training}

Let us consider an MLP using
rectified linear hidden neurons. Then, learning with dropout
is equivalent to training an MLP with auxiliary stochastic
neurons of which each follows Bernoulli distribution with
mean $p=0.5$.
%In this case, when evaluating the activation of each hidden neuron
%$h_j^{\qlay{l}}$ we fix the edge weight $u_j^{\qlay{l}}$
%between this hidden neuron and the auxiliary neuron
%$r_j^{\qlay{l}}$ to the negated \textit{value} of the
%ordinary input
%\footnote{
%This is a \textit{hack} in the sense that we consider
%copying a negated value to $u_j^{\qlay{l}}$ to avoid making
%$u_j^{\qlay{l}}$ actually parameterized by the existing parameters of
%the MLP. This is not necessary for training, as we may
%simply set $u_j^{\qlay{l}}$ to negative infinity to have the
%same effect being explained in this section, but is
%necessary in testing phase.
%}:
%\[
%u_j^{\qlay{l}} = - \alpha_j^{\qlay{l}} =
%-\sum_{i} h_i^{\qlay{l-1}} w_{ij}^{\qlay{l-1}}.
%\]
We fix the weight $u_j^{\qlay{l}}$ of the edge connecting
from the auxiliary stochastic neuron to a hidden neuron  to
negative infinity.  Then, the activation of $h_j^{\qlay{l}}$
is 
\begin{align}
    \label{eq:relu_dropout}
    h_j^{\qlay{l}} = \left\{\begin{array}{l l}
        \max(0, \alpha_j^{\qlay{l}}) & \text{, if }
        r_j^{\qlay{l}} = 0 \\
        0 & \text{, if }
        r_j^{\qlay{l}} = 1 
    \end{array}\right.
\end{align}
This is equivalent to using dropout in training an MLP.

This exact procedure applies to any hidden neuron which has
an activation function that converges to zero in the limit
of negative infinity.  A logistic sigmoid activation
function is one such example. In cases of other types of
activation functions, other ways of fixing the connection
strengths between a hidden neuron and its corresponding
auxiliary neuron are needed.

%Interestingly, this way of viewing the procedure of dropout
%implies that we are training as many MLPs as the number of training samples
%simultaneously. Since the connection weights
%between the hidden neurons and their corresponding auxiliary
%neurons are different among training samples, each training
%sample has a unique MLP with common weights shared with the MLPs
%of the other training samples. However, in the remaining of
%this section we see that
%by considering only the \textit{expected} activation of every
%hidden neuron all those different MLPs collapse into a
%single one.

\subsection{Testing}

When an MLP was trained using a procedure of dropout, it was
proposed in \cite{Hinton2012} that outgoing weights be
halved to compensate for the loss of approximately half of
hidden neurons during training phase.  With a mild
approximation, here we show that this procedure of halving
the outgoing weights corresponds to computing the expected
activation of output neurons over the auxiliary stochastic
neurons.

If we linearly approximate the expectation of the output
neurons, we may push the expectation operator all the way
down to the evaluation of the activation of each hidden
neuron $h_j^{\qlay{l}}$. Because the activation is dropped
to zero with
probability $0.5$, the expected activation of
$h_j^{\qlay{l}}$ becomes, for instance in the case of a
rectified linear hidden neuron in
Eq.~\eqref{eq:relu_dropout},
\begin{align*}
    \E \left[ h_j^{\qlay{l}} \right] = \frac{1}{2} \max\left(0,
    \sum_{i} h_i^{\qlay{l-1}} w_{ij}^{\qlay{l-1}}\right).
\end{align*}
It is clear to see that this is effectively equivalent to
halving the outgoing weights. 
%This procedure effectively
%merges all those different MLPs into a single MLP with the
%halved outgoing weights.

Linear approximation is unnecessary, and computing the
expectation becomes exact, if the output neurons are linear
and there is only a single layer of hidden neurons. This
agrees well with the original formulation of dropout in
\cite{Hinton2012} which formulated the procedure of halving
the outgoing weights as taking a geometric average of
exponentially many neural networks that share parameters.
However, this procedure, in both perspectives, becomes
approximate as the number of nonlinear hidden layers
increases.

From the proposed framework, we can see that, albeit
informally, this procedure of halving the outgoing weights
is well approximated if the activation function of each
hidden neuron can be approximated well linearly. There are
two potential consequences from this. Applying dropout to
hidden neurons below an activation function which may not be well
approximated linearly, such as max-pooling, will not work
well, which has been noticed already by previous work (see,
e.g., \cite{Zeiler2013,Krizhevsky2012}). Secondly, a
piece-wise linear activation function such as the rectified
linear function is well-suited
for using dropout. This agree well with recent finding that
another piece-wise linear activation function called maxout
works well with dropout \cite{Goodfellow2013}. 

By this formulation, we can extend the original dropout by
dropping each hidden neuron with probability $p$ instead of
$0.5$. In that case, in testing time, the outgoing weight
will be multiplied by $1 - p$. Furthermore, this allows us
to use different dropping probabilities for hidden neurons.
If we denote the dropping probability of each hidden neuron
by $p_j^{\qlay{l}}$, this will correspond to multiplying all
outgoing weights $w_{jk}^{\qlay{l}}$ of the $j$-th hidden
neuron in the $l$-th hidden layer with $1 - p_j^{\qlay{l}}$.

\section{Other Special Cases}

In this section, we describe two other popular training
schemes and how they are realized as special cases of the
proposed procedure of adding auxiliary stochastic neurons.
The two training schemes we discuss here are denoising
\cite{Vincent2010,Bishop1995} and semantic hashing
\cite{Salakhutdinov2009s}.

\subsection{Explicit Noise Injection: Denoising Autoencoder}

A denoising autoencoder (DAE) \cite{Vincent2010} is an MLP
that aims to reconstruct a clean sample given an explicitly
corrupted input. The DAE is an obvious special case of the
proposed general framework. In this section, we consider
adding additive white Gaussian noise to each input
component.

A DAE can be constructed from an ordinary autoencoder by
adding an additional hidden layer between the input and the
first hidden layer. The additional layer has as many hidden
neurons as the number of input variables. Each hidden
neuron $\nu_i$ is connected to the $i$-th input component
$x_i$ \textit{only} with weight $1$ and has an auxiliary
stochastic neuron $r_i$ which follows a standard Normal
distribution.

The activation of $\nu_i$ is linear and defined to be
\begin{align*}
    \nu_i = x_i + r_i u_i,
\end{align*}
where $u_i$ is the connection strength between $\nu_i$ and
$r_i$. Each time $\nu_i$ is computed, the activation of
$r_i$ is sampled from a standard Normal distribution. This
is equivalent to explicitly adding additive white Gaussian
noise with variance $r_i^2$.

Once training is over, we can compute the hidden activation
of the original DAE by first computing the expected
activation of $\nu_i$. Since $\E\left[ r_i \right] = 0$, the
activation of $\nu_i$ is  simply a copy of the input $x_i$.
In other words, we can use the learned parameters as if they
were the parameters of an ordinary autoencoder trained
without explicitly adding noise.

By further adding more intermediate hidden layers with
auxiliary stochastic neurons, we can emulate adding multiple
types of noise sequentially to input. For instance, a common
practice of adding white Gaussian noise and dropping a small
portion of input components can be achieved by adding
another intermediate hidden layer that drops some components
randomly, just like dropout described in
Section~\ref{sec:dropout}.

%It is naturally to extend this approach by adding noise to
%\textit{hidden} layers. Although this has not been studied
%extensively yet, it is straightforward to build a neural
%network that implements adding noise to hidden layers under
%the proposed framework. However, this will again require
%approximating the activation of each hidden neuron linearly
%to push the expectation operator inside the activation
%function.

This method of explicitly injecting noise to input is
obviously applicable to a standard MLP that performs
classification \cite{Bishop1995,Raiko2012}. Furthermore,
under the proposed framework this method naturally allows us
to add noise even to hidden neurons, which may work as a
regularization similarly to using dropout. This idea of
adding noise to hidden neurons as well as input variables
has recently been applied to a deep generative stochastic
network in \cite{Bengio2013gsn}.

\subsection{Semantic Hashing}

Semantic hashing was proposed in \cite{Salakhutdinov2009s}
to extract a binary code of a document using a deep
autoencoder with a small sigmoid bottleneck layer. One of
the important details of the training procedure in
\cite{Salakhutdinov2009s} was to add white Gaussian noise to
the input signal to the bottleneck layer to encourage the
activations of the hidden neurons in the bottleneck layer to
be as close to $0$ or $1$ as possible.

This procedure is exactly equivalent to adding an auxiliary
stochastic neuron to each bottleneck hidden neuron. The
activation of each auxiliary stochastic neuron is sampled
from a standard Normal distribution and is multiplied with
the connection strength which corresponds to the variance of
the added noise. Since the connection strength is fixed and
the auxiliary stochastic neuron is independent from the
input or any other neuron, an ordinary backpropagation can
be used without any complication resulting from the stochastic
auxiliary neurons.

Again, once the parameters were learned, one may safely
ignore the added auxiliary stochastic neurons as their means
are zero.

\section{Experiments}

Although this paper focuses on interpreting various recently
proposed training schemes under the proposed framework of
adding auxiliary stochastic neurons. We were able to find
some potentially useful extensions of those existing schemes
by understanding them from this new perspective. One of them
is to extend the usage of dropout by using different
dropping probabilities for hidden neurons, and another is to
inject white Gaussian noise to hidden neurons.

In this section, we present preliminary experiment result
showing the effect of (1) using a separate dropping
probability for each hidden layer and (2) injecting white
Gaussian noise to the \textit{input} of each hidden neuron.

\subsection{Settings}

We trained MLPs with two hidden layers having 2000
rectified linear neurons each on handwritten digit dataset
(MNIST, \cite{Lecun1998}) using either dropout with separate
dropping probabilities for the two hidden layers or
injecting white Gaussian noise. 

To see the effect of choosing separate dropping
probabilities for hidden layers, we trained 100 MLPs with a
dropping probability $p_l$ with the $l$-th hidden layer
randomly selected from the interval $\left[ 0, 1\right]$.
Similarly, 100 MLPs were trained with separate noise
variances for hidden layers, where the exponent $s_l$ of a
noise variance for the $l$-th hidden layer was randomly
chosen from $\left[-5, 0 \right]$.

Before training each MLP, $60,000$ training samples were
randomly split into training and validation sets with ratio
$3:1$.  Learning was early stopped by checking the
prediction error on the validation set, while the maximum
number of epochs was limited to $100$\footnote{Almost all
runs were early-stopped before 100 epochs.}. We used a
recently proposed method, called ADADELTA \cite{Zeiler2012},
to adapt learning rates automatically. Since we fixed the
size of an MLP, this effectively means that there were no
other hyperparameters to tune.

\subsection{Result and Analysis}

\begin{figure}[t]
    \psfrag{p1}[Bl][Bl][1.0][0]{$p_1$}
    \psfrag{p2}[Bl][Bl][1.0][0]{$p_2$}
    \psfrag{v1}[Bl][Bl][1.0][0]{$s_1$}
    \psfrag{v2}[Bl][Bl][1.0][0]{$s_2$}
    \centering
    \begin{minipage}{0.48\textwidth}
        \centering
        \includegraphics[width=1.0\columnwidth]{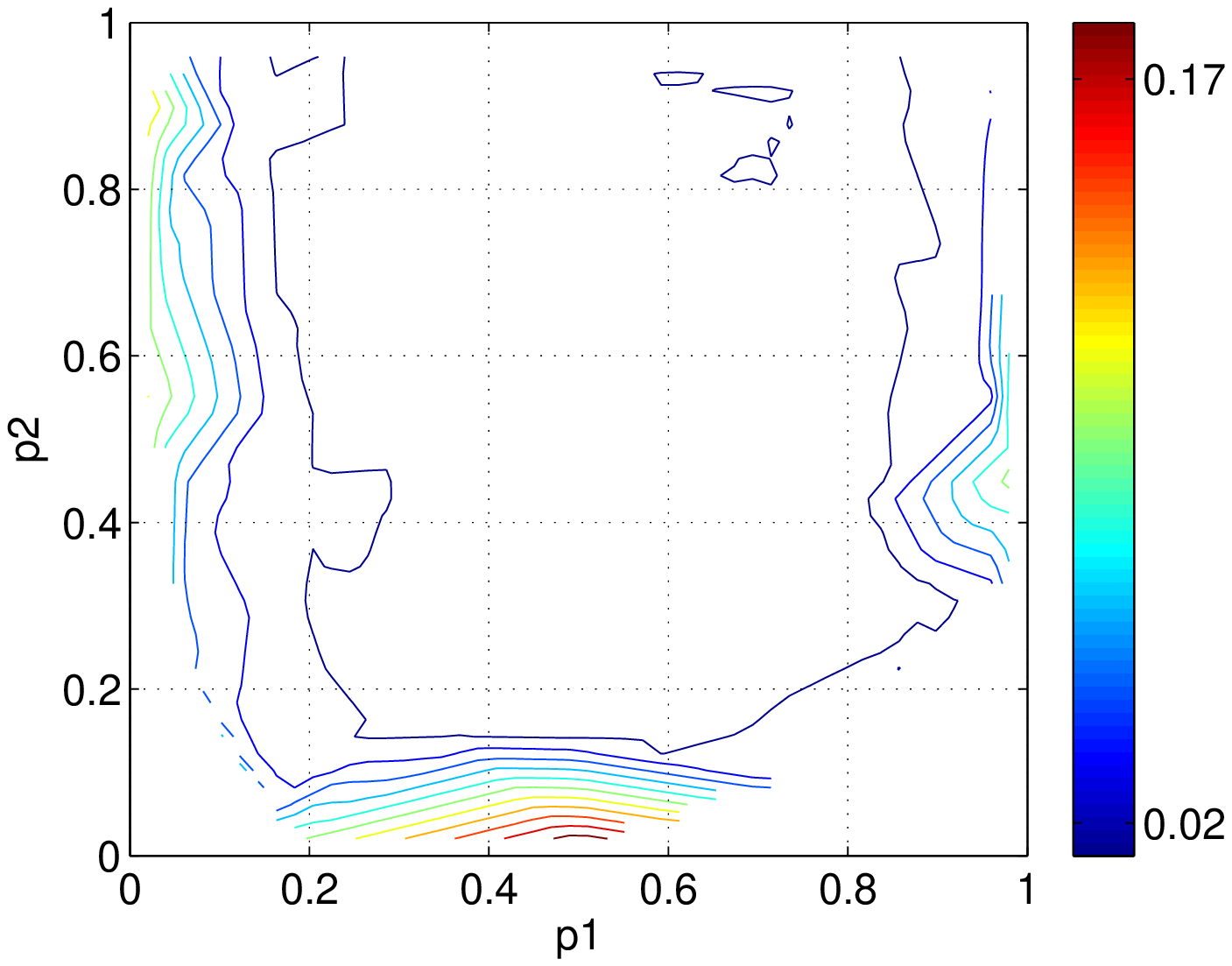}
        \\
        (a)
    \end{minipage}
    \begin{minipage}{0.48\textwidth}
        \centering
        \includegraphics[width=1.0\columnwidth]{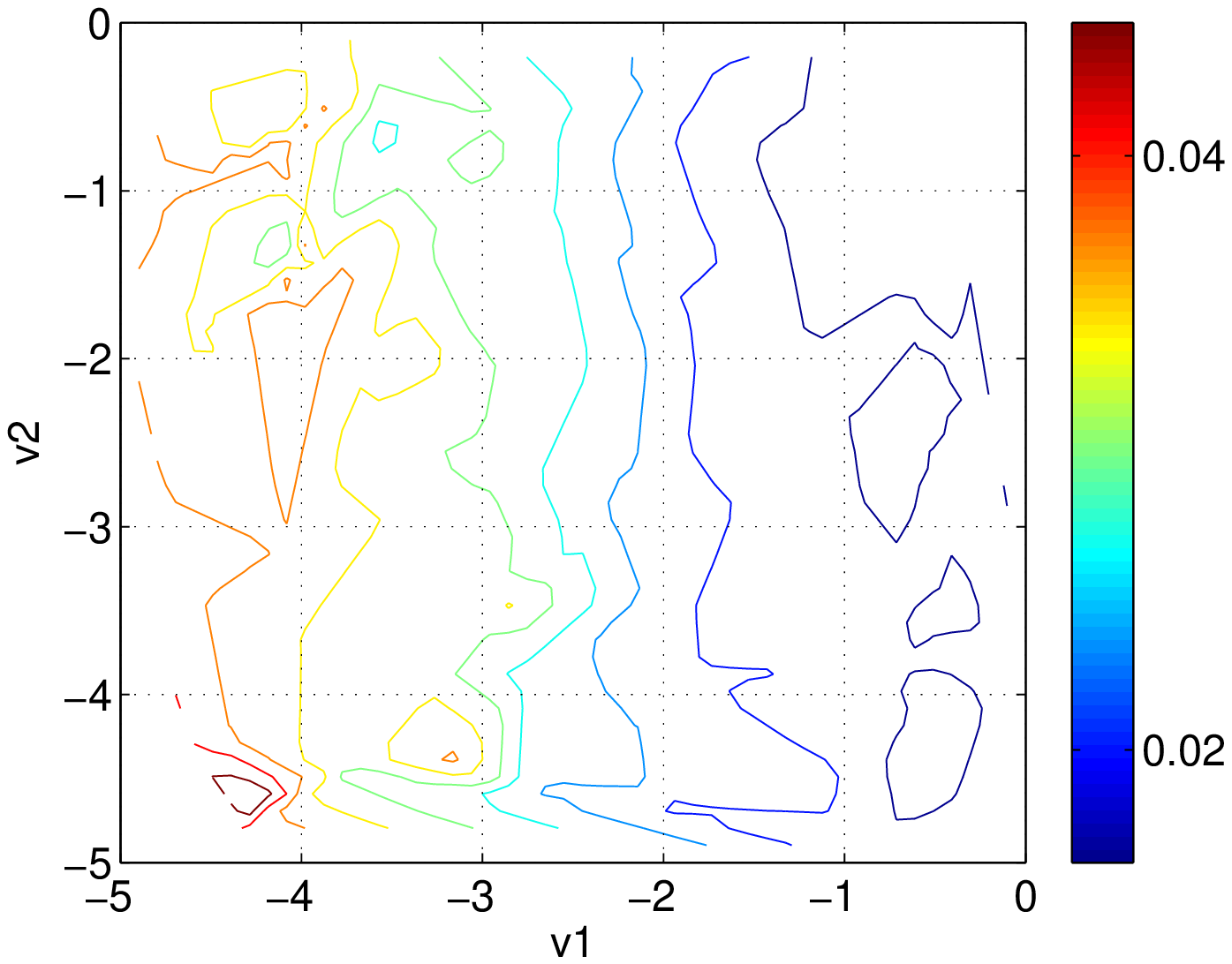}
        \\
        (b)
    \end{minipage}
    \caption{Contour plots of interpolated classification
    errors. (a) Figure obtained by the MLPs
    trained using separate dropping probabilities $p_1$ and
    $p_2$ for two
    hidden layers. (b) Figure obtained by the MLPs trained
    by injecting white Gaussian noise to the inputs to 
    hidden neurons using separate standard deviations $s_1$
    and $s_2$ for two hidden layers. These figures are best
    viewed in color.}
    \label{fig:exp}
    \vspace{-3mm}
\end{figure}

The result for the first experiment tested using
separate dropping probabilities for hidden layers is shown
in Fig.~\ref{fig:exp} (a). Interestingly, it can be observed
that any dropping probability near $0.5$ resulted in
relative good accuracy. However, when any extreme dropping
probability close to either $0$ or $1$ was used for the
first hidden layer ($p_1$), the performance dropped
significantly regardless of the dropping probability of the
second hidden layer ($p_2$). Using too small dropping
probability in the second hidden layer also turned out to
hurt the generalization performance significantly. This
suggests that the original proposal of simply dropping
approximately half of hidden neurons in each hidden layer
from \cite{Hinton2012} is already a good choice.

In Fig.~\ref{fig:exp} (b), the result of the second
experiment is shown. In general, it shows that the
generalization performance of an MLP is highly affected by
the level of noise injected at the first hidden layer, which
is in accordance with the previous research showing that
adding noise to the input improves the classification
accuracy on test samples \cite{Raiko2012}. However, a closer
look at the figure shows that adding noise to the upper
hidden layer helps achieving better generalization
performance (see the upper right corner of the figure).

One important lesson from these preliminary experiments is
that it is possible to achieve better generalization
performance by carefully tuning auxiliary stochastic
neurons. This amounts to, for instance, choosing different
dropping probabilities in the case of dropout and injecting
different levels of Gaussian noise. Further and deeper
investigation using various architectures and datasets is,
however, required.

\section{Discussion}

In this paper, we have described a general method of adding
auxiliary stochastic neurons in a multi-layer perceptron
(MLP). This procedure effectively makes hidden neurons in an
MLP stochastic, but does not require any change to the
standard backpropagation algorithm which is commonly used to
train an MLP.

This proposed method turned out to be a generalization of a
few recently introduced training schemes. For instance,
dropout \cite{Hinton2012} was found to be a special case
having binary auxiliary neurons with connection strengths
dependent on the input signal. A method of explicitly
injecting noise to input neurons \cite{Bishop1995,Raiko2012}
used by, for instance, a denoising autoencoder
\cite{Vincent2010} was found to be an obvious application of
the proposed use of auxiliary stochastic neurons following
standard Normal distribution. Furthermore, we found that a
trick of making the activations of hidden neurons in the
bottleneck layer of an autoencoder used for semantic hashing
\cite{Salakhutdinov2009s} is equivalent to simply adding a
white Gaussian auxiliary stochastic neuron to each hidden
neuron in a bottleneck layer. 

This paper, however, did not attempt to explain why, for
instance, dropout helps achieving better generalization
performance. Training an MLP with dropout under the proposed
framework does not differ greatly from the ordinary way of
training. The only difference is that some randomness is
explicitly defined and injected via auxiliary stochastic
neurons. It is left for future to investigate whether this
simple injection of randomness causes a favorable
performance of an MLP trained with dropout, or there exist
more behind-the-scene explanations.  The same argument
applies to denoising autoencoders and semantic hashing as
well.

One important thing to note is that the proposed method is
\textit{not} equivalent to building an MLP with stochastic
activation functions. It may be possible to find an
equivalent model with auxiliary stochastic neurons, but it
is not guaranteed nor expected that every stochastic MLP can
be emulated by an ordinary MLP augmented with auxiliary
stochastic neurons. However, one advantage of using the
proposed method of adding auxiliary neurons compared to a
true stochastic MLP is that there is no need for modifying
the standard backpropagation or designing a new learning
algorithm (see, e.g., \cite{Bengio2013sto,Tang2013}).
%\cite{Raiko2007}).

By understanding the method of dropout under the proposed
framework, another extension was found, which allows using a
separate dropping probability for each hidden layer.
Similarly, we observed that it is also possible, under the
proposed framework, to inject white Gaussian noise at each
hidden layer instead of injecting only at the input.  In the
experiments, we provided empirical evidence showing that
better generalization performance may be achieved by using
separate dropping probabilities for different hidden layers
in the case of dropout as well as injecting white Gaussian
noise to hidden layers. As the experiments were quite
limited, however, further extensive evaluation is required
in future.

%\small
\bibliographystyle{splncs03}
\bibliography{iconip}

\end{document}